\documentclass[final,11pt,a4paper]{article}

\usepackage{url}
\usepackage[left=25mm,right=25mm,top=25mm,bottom=25mm]{geometry}
\usepackage[normal,small,bf]{caption2}
\usepackage{graphicx}
\usepackage{cite}

\begin{document}


\title{A framework for the interactive resolution of multi-objective vehicle routing problems}
\author{Martin Josef Geiger and Wolf Wenger\footnote{Lehrstuhl f\"{u}r Industriebetriebslehre, Universit\"{a}t
Hohenheim, 70593 Stuttgart, Germany\newline Email:
\texttt{\{mjgeiger|w-wenger\}@uni-hohenheim.de}}}
\date{}
\maketitle

\begin{abstract}
The article presents a framework for the resolution of rich vehicle
routing problems which are difficult to address with standard
optimization techniques. We use local search on the basis on
variable neighborhood search for the construction of the solutions,
but embed the techniques in a flexible framework that allows the
consideration of complex side constraints of the problem such as
time windows, multiple depots, heterogeneous fleets, and, in
particular, multiple optimization criteria. In order to identify a
compromise alternative that meets the requirements of the decision
maker, an interactive procedure is integrated in the resolution of
the problem, allowing the modification of the preference information
articulated by the decision maker. The framework is prototypically
implemented in a computer system. First results of test runs on
multiple depot vehicle routing problems with time windows are
reported.
\end{abstract}

{\small {\bf Keywords:} User-guided search, interactive
optimization, multi-objective optimization, multi depot vehicle
routing problem with time windows, variable neighborhood search.}

\section{Introduction}
The vehicle routing problem (VRP) is one of the classical
optimization problems known from operations research with numerous
applications in real world logistics. In brief, a given set of
customers has to be served with vehicles from a depot such that a
particular criterion is optimized. The most comprehensive model
therefore consists of a complete graph $G = (V, A)$, where $V=\{
v_{0}, v_{1}, \ldots, v_{n} \}$ denotes a set of vertices and $A =
\{(v_{i}, v_{j}) \mid v_{i}, v_{j} \in V, i \neq j\}$ denotes the
connecting arcs. The depot is represented by $v_{0}$, and $m$
vehicles are stationed at this location to service the customers
$v_{i}, \ldots, v_{n}$. Each customer $v_{i}$ demands a nonnegative
quantity $q_{i}$ of goods and service results in a nonnegative
service time $d_{i}$. Traveling on a connecting arc $(v_{i}, v_{j})$
results in a cost $c_{ij}$ or travel time $t_{ij}$. The most basic
vehicle routing problem aims to identify a solutions that serves all
customers, not exceeding the maximum capacity of the vehicles
$Q_{k}$ and their maximum travel time $T_{k}$ while minimizing the
total distances/costs of the routes.

Various extensions have been proposed to this general problem type.
Most of them introduce additional constraints to the problem domain
such as time windows, defining for each customer $v_{i}$ an interval
$[e_{i}, l_{i}]$ of service. While arrival before $e_{i}$ results in
a waiting time, arrival after $l_{i}$ is usually considered to be
infeasible \cite{solomon:1988:article}. In other approaches, the
times windows may be violated, leading to a tardy service at some
customers \cite{taillard:1997:article}.

Some problems introduce multiple depots as opposed to only a single
depot in the classical case. Along with this sometimes comes the
additional decision of open routes, where vehicles do not return to
the place they depart from but to some other depot. Also, different
types of vehicles may be considered, leading to a heterogeneous
fleet in terms of the abilities of the vehicles.

Unfortunately, most problems of this domain are $\mathcal{NP}$-hard.
As a result, heuristics and more recently metaheuristics have been
developed with increasing success
\cite{gendreau:2003:inproceedings}. In order to improve known
results, more and more refined techniques have been proposed that
are able to solve, or at least approximate very closely, a large
number of established benchmark instances. With the increasing
specialization of techniques goes however a decrease in generality
of the resolution approaches.

While the optimality criterion of minimizing the total traveled
distances is the most common, more recent approaches recognize the
vehicle routing problem as a multi-objective optimization problem
\cite{rahoual:2001:inproceedings,geiger:2001:routing,jozefowiez:2002:inproceedings,murata:2005:inproceedings}.
Here, the overall problem lies in identifying a Pareto-optimal
solution $x^{*}$ that is most preferred by a decision maker. As the
relevant objective functions are often of conflicting nature, a
whole set of potential Pareto-optimal solutions exists among which
this choice has to be made.

In the current article, a framework for interactive multi-objective
vehicle routing is presented that aims to address two critical
issues: (i) the necessary generality of resolution approaches when
trying to solve a range of problems of different characteristics,
and (ii) the integration of multiple objectives in the resolution
process.

\section{A framework for interactive multi-objective vehicle
routing}

Independent from the precise characteristics of the particular VRP,
two types of decisions have to be made when solving the problem.
\begin{enumerate}
\item Assignment of customers to vehicles (clustering).
\item Construction of a route for a given set of customers (sequencing).
\end{enumerate}

\begin{figure}[!ht]
\begin{center}
\includegraphics{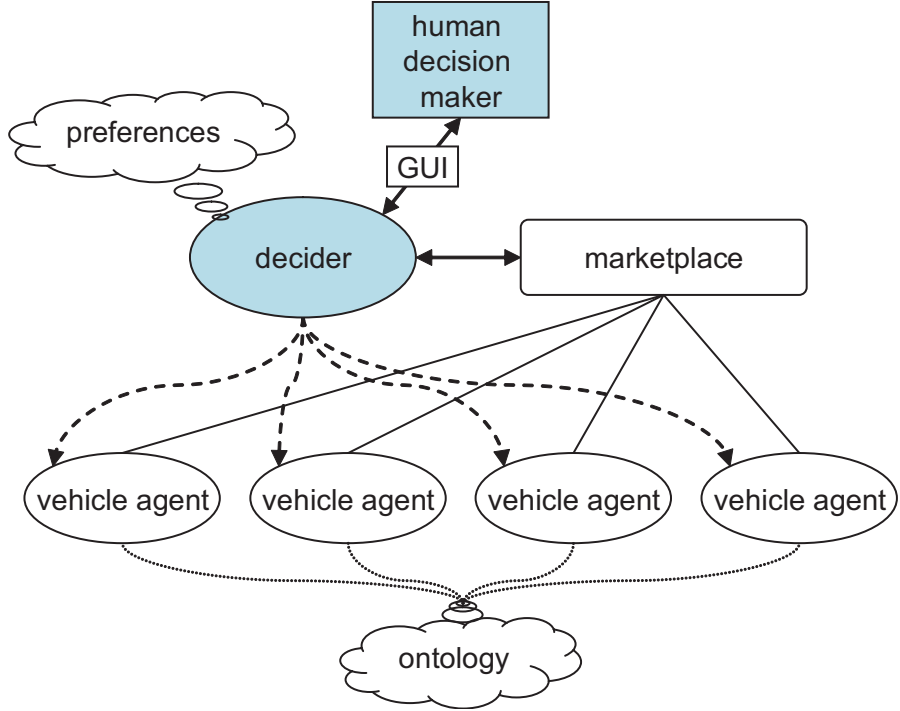}
\caption{\label{fig:framework}Sketch of the framework}
\end{center}
\end{figure}

It is well-known that both types of decisions influence each other
to a considerable extent. The here presented framework therefore
proposes the use of a set of elements to handle this issue with
upmost generality. Figure~\ref{fig:framework} gives an overview
about the elements used.

\begin{itemize}
\item[--] The {\em marketplace} represents the element where orders are
offered for transportation.
\item[--] {\em Vehicle agents} place bids for orders on the marketplace.
These bids take into consideration the current routes of the
vehicles and the potential change when integrating an additional
order.
\item[--] An {\em ontology} describes the precise properties of the
vehicles such as their capacity, availability, current location,
etc. This easily allows the consideration of different types of
vehicles.
\item[--] A {\em decider} communicates with the human decision maker via
a graphical user interface (GUI) and stores his/her individual
preferences. The decider also assigns orders to vehicles, taking
into consideration the bids placed for the specific orders.
\end{itemize}

A solutions is constructed by placing the orders on the marketplace,
collecting bids from the vehicle agents, and assigning orders to
vehicles while constantly updating the bids. Route construction by
the vehicle agents is done in parallel using local search heuristics
so that a route can be identified that maximizes the preferences of
the decision maker.

\section{Implementation and preliminary experiments}
The framework has been prototypically implemented in a computer
system. In the first experiments, two objective functions are
considered, the total traveled distances $DIST$ and the total
tardiness $TARDY$ caused by vehicles arriving after the upper bound
$l_{i}$ of the time window.

The preferences of the decision maker are represented introducing a
weighted sum of both objective functions. Using the relative
importance of the distances $w_{DIST}$, the overall utility UTILITY
of a particular solution can be computed as given in
Expression~\ref{eqn:utility}.

\begin{equation}
\label{eqn:utility} UTILITY = w_{DIST}\,\, DIST + (1 - w_{DIST})\,\,
TARDY
\end{equation}

The vehicle agents are able to modify the sequence of their orders
using four different local search neighborhoods.
\begin{itemize}
\item[--] Inverting the sequence of the orders between positions $p_{1}$ and
$p_{2}$. While this may be beneficial with respect to the distances,
it may pose a problem for the time windows as usually orders are
served in the sequence of their time windows.
\item[--] Exchanging the positions $p_{1}$ and $p_{2}$ of two orders.
\item[--] Moving an order from position $p_{1}$ and reinserting it at
position $p_{2}$, $p_{1} < p_{2}$ (forward shift).
\item[--] Moving an order from position $p_{1}$ and reinserting it at
position $p_{2}$, $p_{1} > p_{2}$ (backward shift).
\end{itemize}
In each step of the local search procedure, a neighborhood is
randomly picked from the set of neighborhoods and a move is computed
and accepted given an improvement.

Bids for orders on the marketplace are generated by the vehicle
agents, taking into consideration all possible insertion points in
the current route. The sum of the weighted increase in distance DIST
and tardiness TARDY gives the prize for the order.

The decider assigns orders to vehicles such that the maximum regret
when {\em not} assigning the order to a particular vehicle, and
therefore having to assign it to some other vehicle, is minimized.
It also analyzes the progress of the improvement procedures. Given
no improvement for a certain number of iterations, the decider
forces the vehicle agents to place back orders on the market such
that they may be reallocated.

The optimization framework has been tested on a benchmark instance
taken from \cite{cordeau:2001:article}. The instance comprises 48
customers that have to be served from 4 depots, each of which
possesses two vehicles.

\begin{figure}[!ht]
\begin{center}
\includegraphics{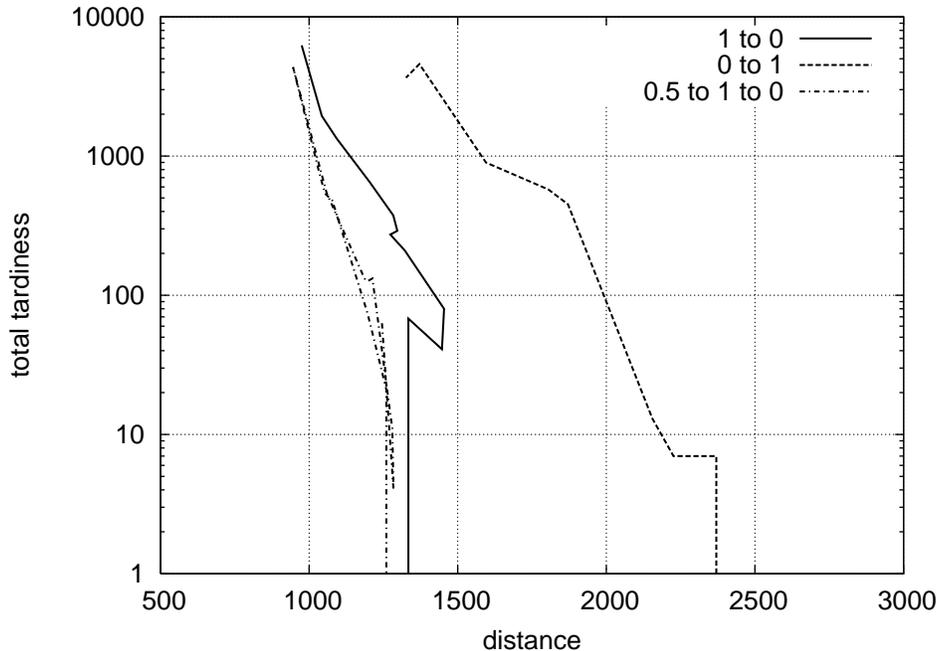}
\caption{\label{fig:testrun}Results of the test runs}
\end{center}
\end{figure}

We simulated a decision maker changing the relative importance
$w_{DIST}$ during the optimization procedure. First, a decision
maker starting with a $w_{DIST} = 1$ and successively decreasing it
to 0, second a decision maker starting with a $w_{DIST} = 0$ and
increasing it to 1, and third a decision maker starting with a
$w_{DIST} = 0.5$, increasing it to 1 and decreasing it again to 0.
Between adjusting the values of $w_{DIST}$ in steps of 0.1, enough
time for computations has been given to the system to allow a
convergence to (at least) a local optimum. Figure~\ref{fig:testrun}
plots the results obtained during the test runs.

The first decision maker starts with $DIST = 975$, $TARDY = 6246$
and moves to $DIST = 1412$, $TARDY = 0$ while the second starts with
$DIST = 2953$, $TARDY = 0$ and moves to $DIST = 1326$, $TARDY =
3654$. Clearly, the first strategy outperforms the second. While an
initial value of $w_{DIST} = 0$ allows the identification of a
solution with zero tardiness, it tends to construct routes that,
when decreasing the relative importance of the tardiness, turn out
to be hard to adapt. In comparison to the strategy starting with a
$w_{DIST} = 1$, the clustering of orders turns out the be
prohibitive for a later improvement.

When comparing the third strategy of starting with a $w_{DIST} =
0.5$, it becomes obvious that this outperforms both other ways of
interacting with the system. Here, the solutions start with $DIST =
1245$, $TARDY = 63$, go to $DIST = 946$, $TARDY = 4342$, and finally
to $DIST = 1335$, $TARDY = 0$. Apparently, starting with a
compromise solution is beneficial even for both extreme values of
$DIST$ and $TARDY$.

\section{Summary and further development}
A framework for the interactive resolution of multi-objective
vehicle routing problems has been presented. The concept has been
prototypically implemented in a computer system. Preliminary results
on a benchmark instance have been reported.

\begin{figure}[!ht]
\begin{center}
\includegraphics[width=8cm]{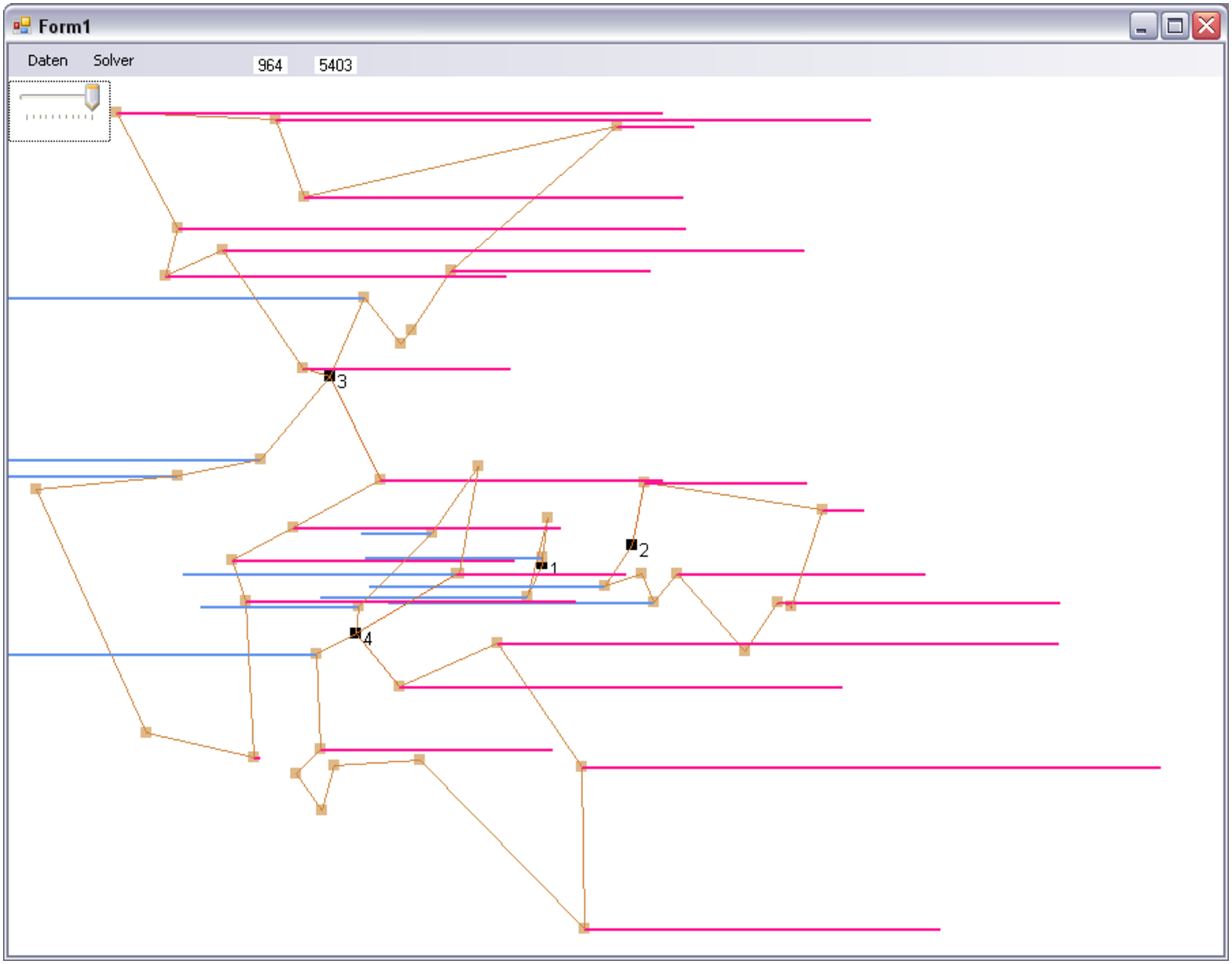}\includegraphics[width=8cm]{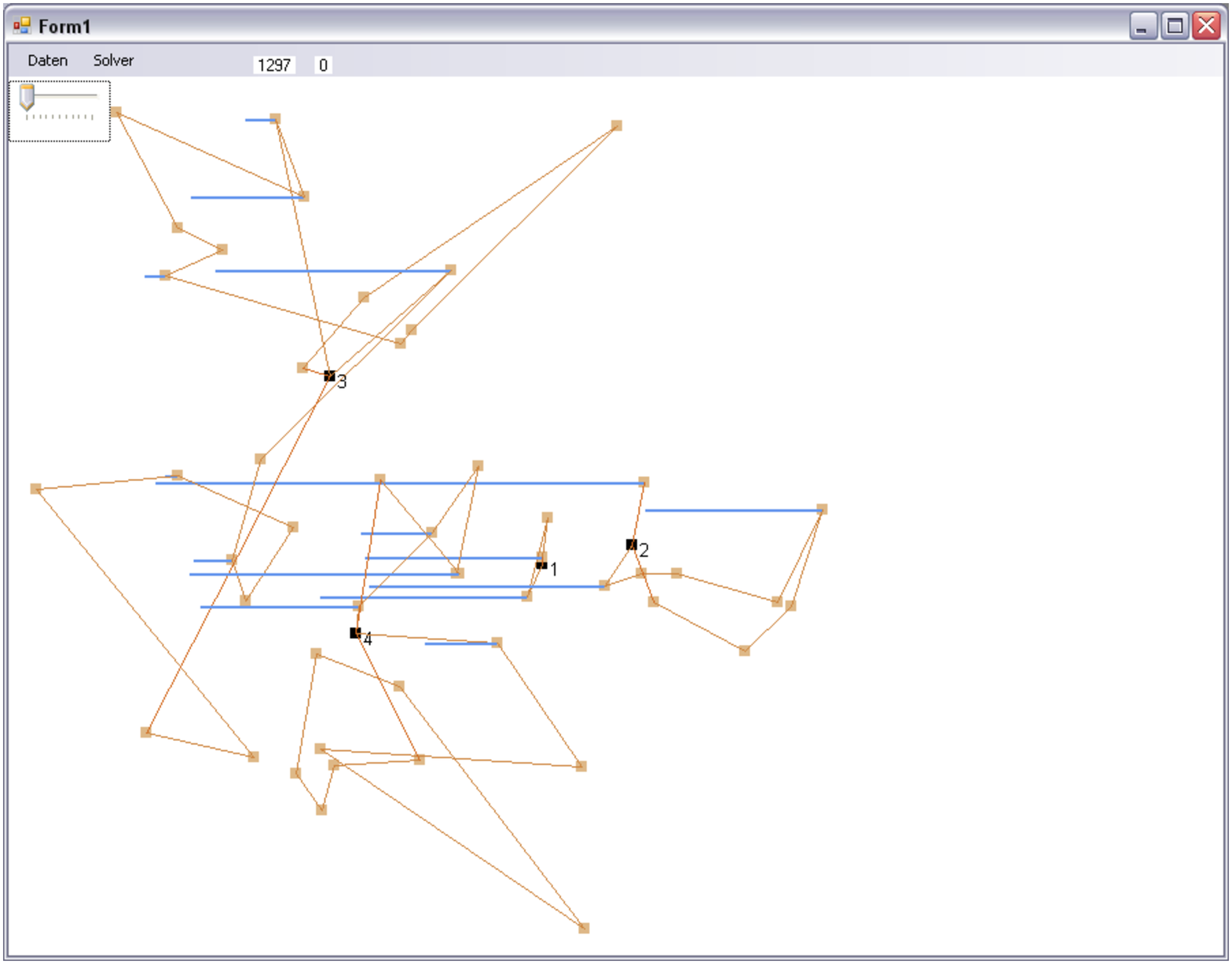}
\caption{\label{fig:interface}Two screenshots of the graphical user
interface. On the left, a short solution with high tardiness, on the
right, a solution with low tardiness but long traveling distances.}
\end{center}
\end{figure}

First investigations indicate that the concept may successfully
solve vehicle routing problems under multiple objectives and complex
side constraints. In this context, an interaction with the system is
provided by a graphical user interface. The relative importance of
the objective functions can be modified by means of a slider bar,
resulting in different solutions which are computed in real time by
the system, therefore providing an immediate feedback to the user.
Figure~\ref{fig:interface} shows two extreme solutions that have
been interactively obtained by the system.

Future developments are manifold. First, other ways of representing
preferences than a weighted sum approach may be beneficial to
investigate. While the comparable easy interaction with the GUI by
means of a slider bar enables the user to directly change the
relative importance of the objective functions, it prohibits the
definition of more complex preference information, e.\,g.\ involving
aspiration levels.

Second, different and improved ways of implementing the market
mechanism have to be investigated. First results indicate that the
quality of the solutions is biased with respect to the initial
setting of the relative importance of the optimality criteria. It
appears as if more complex reallocations of orders between vehicles
are needed to address this issue.

Finally, more investigations on benchmark instances will be carried
out. Apart from test cases known from literature we aim to address
particularly problems with unusual, complex side constraints and
multiple objectives. An additional use of the system will be the
resolution of dynamic VRPs. The market mechanism provides a platform
for the matching of offers to vehicles without the immediate need of
accepting them, yet still obtaining feasible solutions and gathering
a prize for acceptance of offers which may be reported back to the
customer.

\end{document}